\documentclass[a4paper, 10pt, conference]{ieeeconf}      

\IEEEoverridecommandlockouts                              
\usepackage[utf8]{inputenc}
\usepackage[T1]{fontenc}
\usepackage{graphicx}
\usepackage{lipsum,graphicx,multicol}
\usepackage{float}

\usepackage{subcaption}
\usepackage{upgreek}
\graphicspath{ {./images/} }
\usepackage{hyperref}
\hypersetup{
    colorlinks=true,
    linkcolor=black,
    filecolor=magenta,      
    urlcolor=blue,
}

\title{\LARGE \bf
Optimising the selection of samples for robust lidar camera calibration}

\author{Darren Tsai$^{1}$, Stewart Worrall$^{1}$, Mao Shan$^{1}$, Anton Lohr$^{2}$, and Eduardo Nebot$^{1}$
\thanks{$^{1}$D. Tsai, S.Worrall, E.Nebot are with the Australian Centre for Field Robotics (ACFR) at the University of Sydney (NSW, Australia). E-mails: {\tt\small{\{d.tsai, s.worrall,
e.nebot}\}@acfr.usyd.edu.au.} \newline
$^{2}$ Anton Lohr is with Baraja Pty Ltd. Email: \tt\small{anton.lohr@baraja.com}}%
}
\begin{document}

\maketitle
\thispagestyle{empty}
\pagestyle{empty}

\begin{abstract}

We propose a robust calibration pipeline that optimises the selection of calibration samples for the estimation of calibration parameters that fit the entire scene. We minimise user error by automating the data selection process according to a metric, called Variability of Quality (VOQ) that gives a score to each calibration set of samples. We show that this VOQ score is correlated with the estimated calibration parameter's ability to generalise well to the entire scene, thereby overcoming the overfitting problems of existing calibration algorithms. Our approach has the benefits of simplifying the calibration process for practitioners of any calibration expertise level and providing an objective measure of the quality for our calibration pipeline's input and output data. We additionally use a novel method of assessing the accuracy of the calibration parameters. It involves computing reprojection errors for the entire scene to ensure that the parameters are well fitted to all features in the scene. Our proposed calibration pipeline takes 90s, and obtains an average reprojection error of 1-1.2cm, with standard deviation of 0.4-0.5cm over 46 poses evenly distributed in a scene. This process has been validated by experimentation on a high resolution, software definable lidar, Baraja Spectrum-Scan™; and a low, fixed resolution lidar, Velodyne VLP-16. We have shown that despite the vast differences in lidar technologies, our proposed approach manages to estimate robust calibration parameters for both. Our code and data set used for this paper are made available at \url{https://gitlab.acfr.usyd.edu.au/its/cam_lidar_calibration}.

\end{abstract}

\section{INTRODUCTION}
Autonomous vehicles require quick and accurate perception of their surroundings to make informed, real-time decisions. Cameras are excellent at providing texture and colour information which have shown to be important features for classification. However, cameras often struggle to give meaningful information when faced with non-favourable illumination conditions such as dim or bright lighting conditions. Light Detection and Ranging (lidar) systems, on the other hand, are more robust to illumination variance and are able to effortlessly obtain accurate depth information from the scene. Combining 3D depth information with 2D colour and texture information allows autonomous vehicles to overcome the implicit limitations of these sensor modalities for more robust accurate sensing capabilities. To fully leverage the benefits of both camera and lidar, it is essential to have an accurate extrinsic calibration to obtain the rotation and translation parameters between these two sensor frames. 

\begin{figure}[thpb]
  \centering
  \includegraphics[width=\columnwidth]{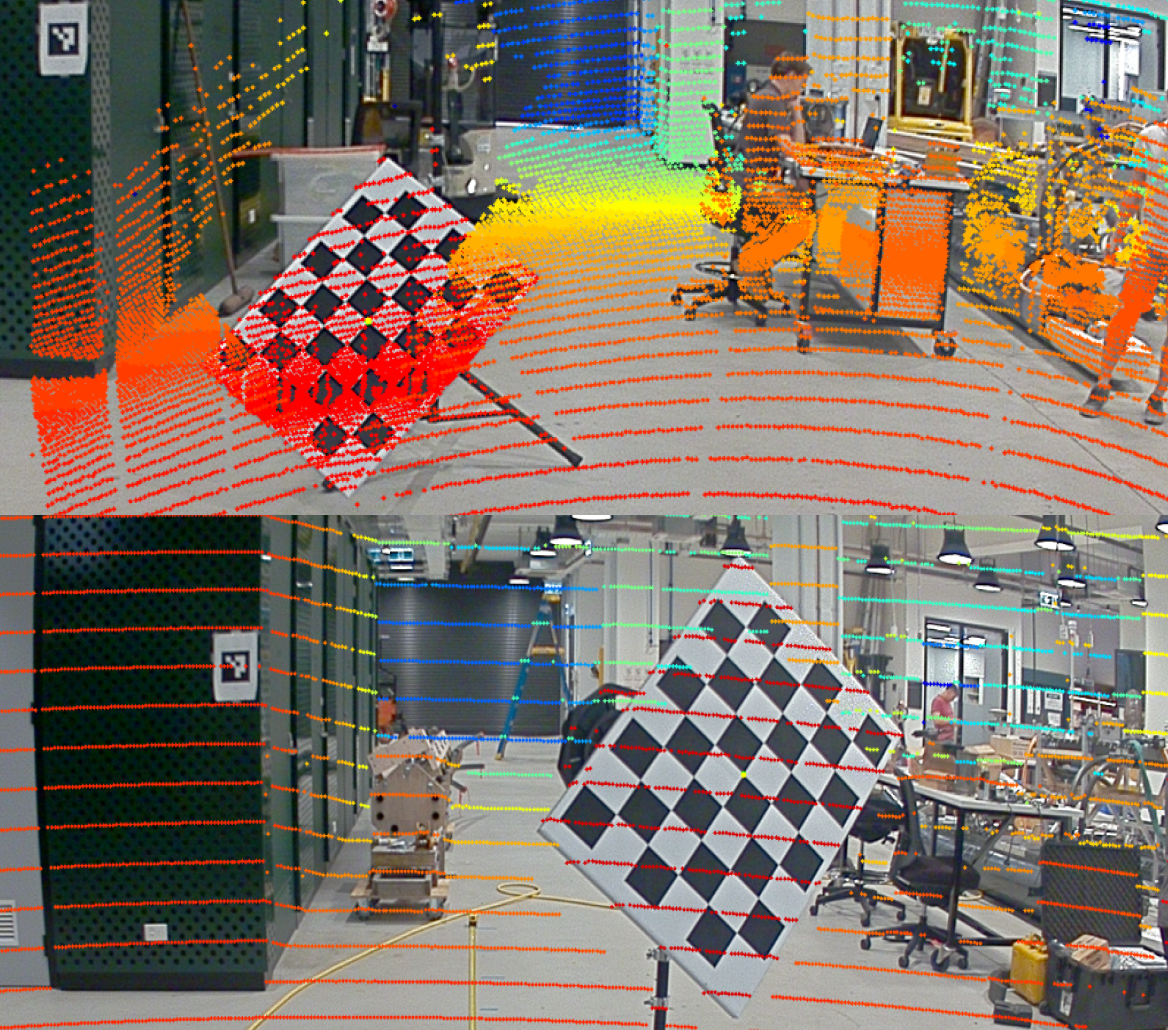}
  \begin{subfigure}[t]{\columnwidth}
        \centering
        \includegraphics[width=\textwidth]{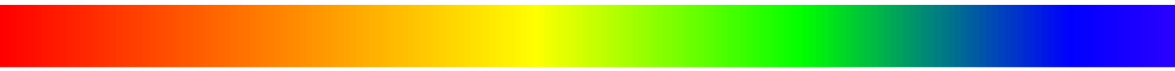}
        \label{fig:colorbar}
    \end{subfigure}%
  \caption{\small Calibration result of Baraja Spectrum-Scan™ (top) and Velodyne VLP-16 (bottom) with our method. Projection is shown for the whole scene, to illustrate that our estimated calibration parameters are fitted to the scene. Points are coloured based on distance to the lidar, red for points closer than 3m and blue for points around 20m away. The chessboard for the Baraja Spectrum-Scan™ (top) has ground points projected on the board due to a large difference in perspective of the Baraja Spectrum-Scan™ and camera. The positioning of the sensors are shown in Fig. \ref{fig:sensor_setup}}
  \label{fig:calib_showcase}
\end{figure}

\begin{figure*}[thbp]
    \centering
    \begin{subfigure}[t]{0.188\textwidth}
        \centering
        \includegraphics[width=\textwidth]{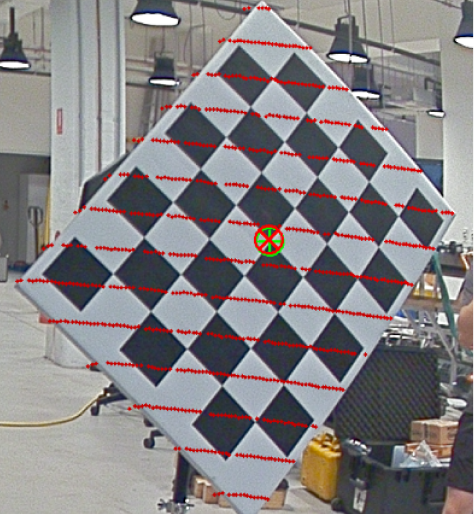}
        \caption{Overfitted (board only)}
        \label{fig:overfit_a}
    \end{subfigure}%
    \hfill
    \begin{subfigure}[t]{0.188\textwidth}
        \centering
        \includegraphics[width=\textwidth]{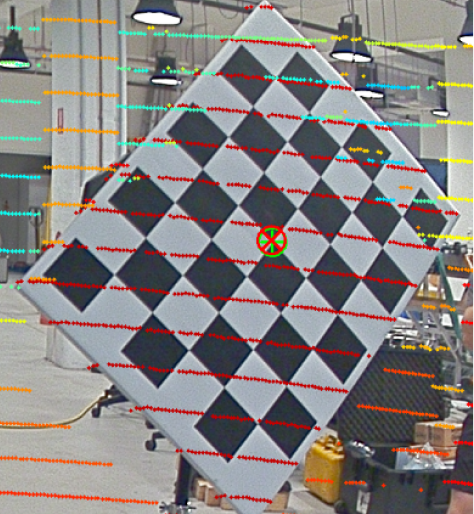}
        \caption{Overfitted (scene)}
        \label{fig:overfit_b}
    \end{subfigure}
    \hfill
    \begin{subfigure}[t]{0.188\textwidth}
        \centering
        \includegraphics[width=\textwidth]{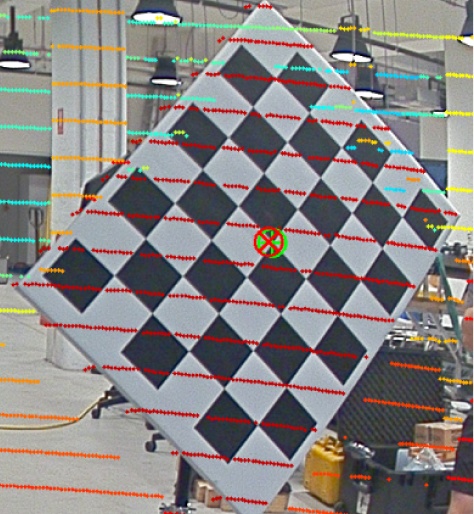}
        \caption{Well-fitted scene}
        \label{fig:notoverfitted}
    \end{subfigure}
    \hfill
    \begin{subfigure}[t]{0.188\textwidth}
        \centering
        \includegraphics[width=\textwidth]{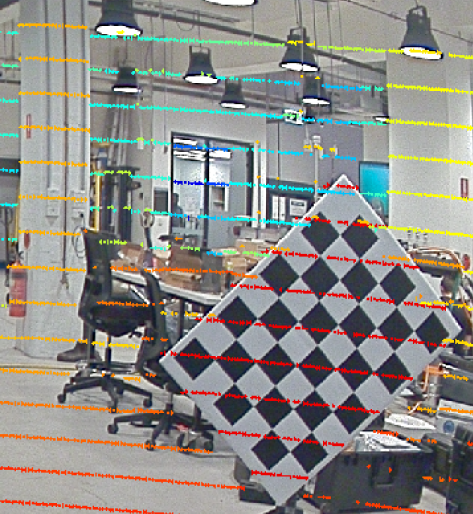}
        \caption{3 poses in a set}
        \label{fig:k3}
    \end{subfigure}
    \hfill
    \begin{subfigure}[t]{0.188\textwidth}
        \centering
        \includegraphics[width=\textwidth]{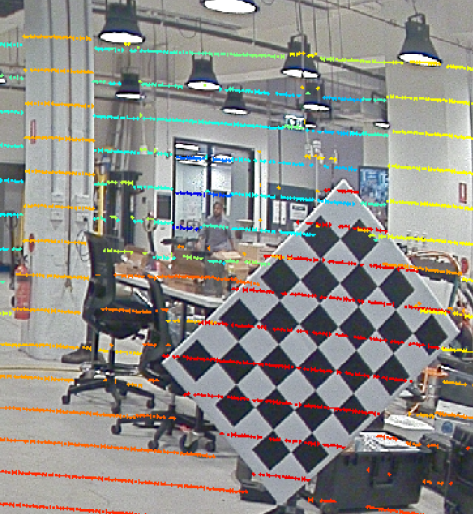}
        \caption{9 poses in a set}
        \label{fig:k9}
    \end{subfigure}
    \caption{\small Many existing methods \cite{Zhou2018,Toth2020, Pusztai2017,Kim2019} often show the effectiveness of their approach by presenting a well-fitted target with low reprojection error (a), but not the whole scene (b). Both (a) and (b) have the same calibration parameters. The centre of the red tilted crosshair is the lidar’s reprojected board centre, and green is the camera’s board centre. The reprojection error in (a) and (b) is 1.8 pixels, which is good, however the calibration parameters don't fit the pillar well at the background because the algorithm only focuses on optimising the chessboard fit. (c) has better overall alignment of scene features than (b) despite having 2.7 pixel error vs 1.7 respectively, showing that same target pose can be fitted with multiple different calibration parameters. This means that there is quite a bit of uncertainty in the accuracy of the fitting of a target in calibration. (d) and (e) compare the calibration result of 3 and 9 poses respectively, showing that an equally good calibration can be achieved on both.}
    \label{fig:calibration_issues}
\end{figure*}

In practical applications, having an accurate calibration enables the projection of lidar points into an image to create a depth-image, or the converse, to project colour information into the pointcloud to create a colourised pointcloud. This has significant implications as it allows us to not only make better-informed decisions but also to fully exploit the capabilities of each sensor in their respective domains. An example of this is the projection of semantic information from images labelled by a CNN to create a labelled pointcloud \cite{Berrio2018, Zhong2017}, or using 2D object bounding boxes to generate 3D frustums as region proposals for 3D object detection \cite{Qi2017, Wang2019}.

The fundamental theoretical problem in calibration is the computation of the transform between two frames (camera and lidar in our case), which is solved in theory but challenging, and often underestimated in its application. The primary challenge is attaining robust and accurate estimation of the calibration in practical applications which is often affected by human, environmental and sensor errors. Existing calibration works can be categorised into two categories: (1) Target-based, which operates in a controlled, and often indoor environment to minimise the influence of environmental factors; and (2) Targetless, which uses features from the environment and aims to remove the human element of the calibration process. 

Whilst targetless approaches are convenient \cite{Scaramuzza2007, Pandey2012, Levinson2013, Irie2016}, existing methods often require a good initial calibration and are not as accurate due to the noise and inconsistencies of feature association from the natural scene \cite{Liao2018, Pusztai2017}. Target-based approaches on the other hand, while more tedious and time-consuming, are able to provide more accurate calibration without requiring an initial estimate. In using either of these approaches, it is therefore often a trade-off between convenience and accuracy. Accuracy in calibration, however, is without a doubt, of higher importance as the calibration result forms the foundation of which many multi-modal perception algorithms build upon. Therefore this paper focuses on improving target-based methods to be more accurate and robust to human and sensor error.

Many successful target-based methods have been proposed for the extrinsic calibration of camera and lidar, however, in practice, their calibration process is often rife with trial and error of finding the best poses for the perfect calibration. Extrinsic calibration parameters are very sensitive, as small errors in rotation or translation estimation can drastically affect the usability of the calibration result \cite{Huang2019}. This sensitivity can most commonly be attributed to noise in the feature extraction process due to errors in sensor measurement which leads to uncertainty in plane-fitting \cite{Park2014, Verma2019, Huang2019} or line-fitting of edges \cite{Park2014, Verma2019}. 

It is challenging to find the perfect calibration as there is no ground truth. To assess the quality of the calibration result, the reprojection error on the alignment of a target is often computed. However, while these methods craft clever cost functions to optimise the alignment of the target, it often results in an overfitted target and an underfitted scene as shown in Fig. \ref{fig:overfit_a} and Fig. \ref{fig:overfit_b}. This is because while the target might be fitted at closer ranges, the errors of estimated parameters grow with distance, becoming more noticeable on objects that are farther away. Many existing works \cite{Zhou2018,Toth2020, Pusztai2017,Kim2019} may often only show Fig. \ref{fig:overfit_a} and report a low reprojection error for it, however this is misleading for those who attempt to use such methods, thinking that they would get a good calibration for the entire scene. Furthermore, there are a range of possible parameters that can fit the target well, but not all of them fit well to the scene as can be seen when comparing Fig. \ref{fig:overfit_b} and Fig. \ref{fig:notoverfitted}.

The target-based calibration pipeline commonly includes 3 sections: data collection, feature extraction and the calibration algorithm. Many authors typically focus on the feature extraction and algorithmic aspect of calibration, however, less attention has been given to the crucial data collection aspect. The importance of good data is commonly stressed in many machine learning algorithms, but not often so in extrinsic calibration methods. In this work, we demonstrate that the estimated calibration parameter's proper alignment of scene features and their uncertainty are highly influenced by the quality of data that is fed into the calibration algorithm. We adopt an existing target-based method \cite{Verma2019} for calibration, as we believe that the existing algorithms are effective, but suffer from user error due to the challenges of collecting good quality samples. 

\begin{figure*}[thpb]
  \centering
  \includegraphics[width=\textwidth]{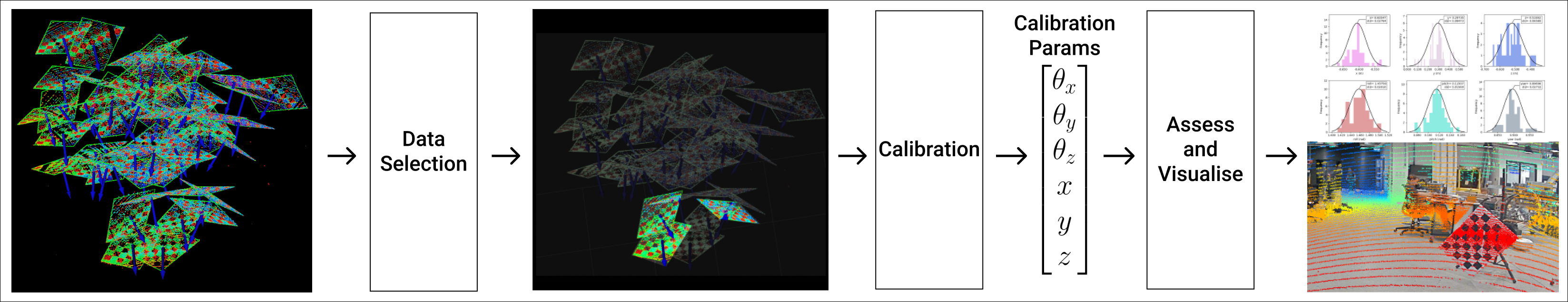}
  \caption{\small Our proposed calibration pipeline. The entire pipeline (excluding the manual data collection phase) takes around 90s to assess the data and estimate calibration parameters. }
  \label{fig:pipeline}
\end{figure*}

The motivation for our work is to provide a calibration tool that fits well to the scene and can be used in an out-of-the-box fashion by any user regardless of calibration expertise. Our method allows the user to simply record many poses with as much variation as possible, and let our algorithm automatically select the best poses to calibrate with. In effect, our work seeks to quantify the calibration expert's knowledge of pose selection in order to make the calibration process less arduous, less time-consuming, and easily accessible. Our contributions can be summarised in the following.
\begin{enumerate}
    \item We propose the Variability of Quality (VOQ) metric to assess the calibration data, and give an indication of the generalisation ability and uncertainty of the estimated calibration parameters. The features used in this metric are common features used in most other target-based methods.
    \item We argue that calibration quality is not about how well a target is fitted, but rather, how well the calibration parameters fit the entire scene. To assess this we propose a robust evaluation of our proposed calibration pipeline that computes the mean and standard deviation of the reprojection error of poses that are evenly spread out in the calibration area. 
    \item The code and data set used in this paper are made available (link in the abstract).
\end{enumerate}

\section{RELATED WORK}

Target-based methods estimate calibration parameters by simultaneously observing the same calibration target from both image and pointcloud domains for feature alignment. The types of targets can be categorised into (1) Planar, such as a chessboard \cite{Zhang2004, Unnikrishnan2005, Pandey2010, Zhou2012, GeigerMoosmann2012, Verma2019}, triangles\cite{Park2014}, and fiducial markers \cite{Dhall2017}; and (2) Non-planar types such as trihedrons \cite{Pusztai2017}, and spheres \cite{Toth2020}. Of the existing targets, the chessboard is the most popular as its pattern provides distinct edges which make it easily identifiable in the image domain. Additionally, the chessboard is straightforward to prepare and easily accessible as it is widely used in intrinsic camera calibration, a pre-requisite of any extrinsic calibration process.

In existing target-based methods, common features extracted from a target are the plane of the target \cite{Zhou2018, Verma2019, Pandey2010, Pusztai2017}, centres and vertices \cite{Verma2019, Alismail2012, Pusztai2017}, and edge lines \cite{Zhou2018}. A popular choice for fitting a line or plane in the lidar domain is RANdom SAmple Consensus (RANSAC) \cite{Verma2019, Pandey2010, Alismail2012, Park2014, Pusztai2017} as it is robust in managing the outliers from noisy lidar returns. Once the plane of the target is obtained, the normal to the plane can be used to represent the plane, often for estimation of the rotation parameters. The corners of the target, and the centre point, can also be obtained in a similar fashion by finding the intersection of the lines that are fitted to the edges of the target.

When target features are extracted, the transform parameters between the lidar and camera can be estimated by computing linear systems of equations and optimised using techniques such as Levenberg-Marquardt \cite{Levenberg1944, Zhang2004, Zhou2018, Pandey2010, Alismail2012, Park2014}, Gauss-Newton \cite{GeigerMoosmann2012}, Iterative Closest Point (ICP) \cite{Alismail2012, GeigerMoosmann2012,Dhall2017} and Genetic Algorithm \cite{Verma2019}. The designed cost functions are usually variants of a least-squares error \cite{Zhang2004, Zhou2012, Pusztai2017, Verma2019} for feature alignment and the reprojection error \cite{Zhang2004, Pandey2010, Verma2019, GeigerMoosmann2012}.

There is minimal literature on the selection of the target poses, and even within target-based methods, such explanations are usually brief. Many target-based works \cite{Pandey2010, Zhou2012, Verma2019, Kim2019} have pointed out that a minimum of 3 poses is required to fully constrain the six degrees of freedom pose, and that there should be a good variation of the poses. However, there is little explanation to aid the user in selecting the ideal 3 poses for the perfect calibration. Other methods simply place lots of chessboards in the scene \cite{GeigerMoosmann2012} which is not always practical, while others do not go into much detail about the choice of poses. \cite{Alismail2012, Park2014, Huang2019}

The work most related to ours is by Zhou et al. \cite{Zhou2012} who observed that the matrix of normals, either of the camera $\mathbf{N_C}$  or lidar $\mathbf{N_L}$, must be full rank, otherwise the configuration has redundant poses and leads to a large estimation error in the results. A weight was assigned to each pose in the dataset based on corner reprojection error, residual error of the fitted plane, and uncertainty of the plane normal. They found that smaller weights were generally assigned to poses that were farther away from the camera, and those that were approximately parallel to the principal plane of the camera. To address the problem of degenerate poses, Zhou et al. \cite{Zhou2018}, in a later work, proposed the estimation of calibration parameters with only a single pose by using the line and plane correspondences of a single chessboard pose to constrain the problem.

\begin{figure*}[thbp]
  \centering
  \vspace{2mm}
  \includegraphics[width=\textwidth]{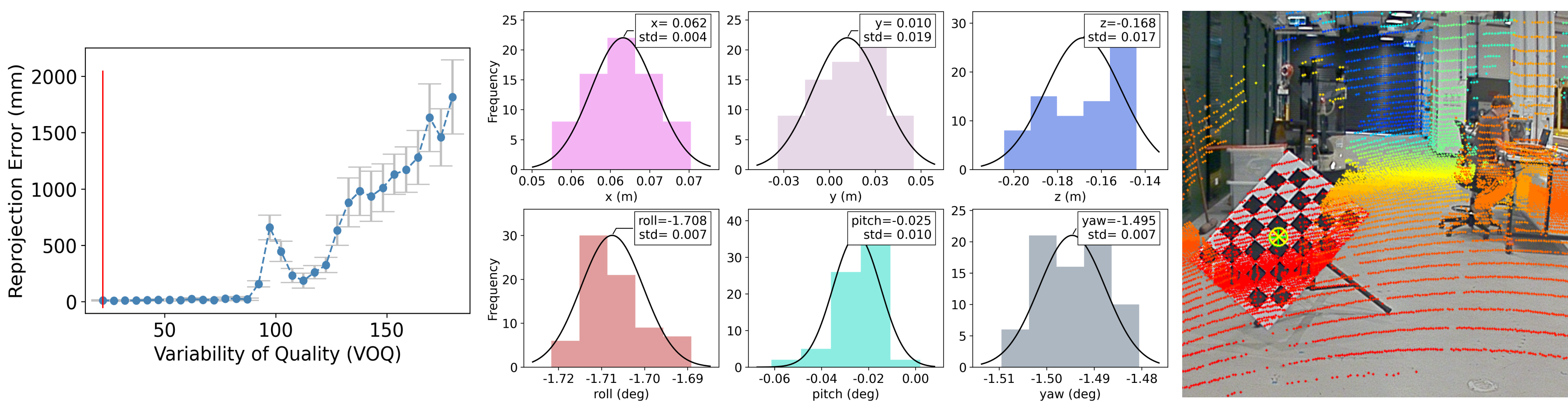}
  \caption{\small We compute the average VOQ of the 50 chosen sets (red line on the left plot), and model a gaussian (middle) to the results generated from these sets to obtain our estimated parameters with uncertainty. Despite a low VOQ, results can vary as each pose sets are specifically fitted to their range. Therefore, we take the average of 50 sets for robustness.}
  \label{fig:pipeline_output}
\end{figure*}

\section{Methodology}

We adopt the calibration method in \cite{Verma2019} which extracts the centre point and plane normal $\mathbf{n_L}$ of a chessboard target. The rotation matrix $\mathbf{R^L_C}$ rotates a matrix of camera normals $\mathbf{N_C}=[\mathbf{n^0_C},\mathbf{n^1_C},\mathbf{n^2_C}]^T$ to align with the corresponding lidar normals $\mathbf{N_L}=[\mathbf{n^0_L},\mathbf{n^1_L},\mathbf{n^2_L}]^T$ as shown in Eq. (\ref{eq:fullyconstrainR}), which we can solve for, given that we know $\mathbf{N_C}$ and $\mathbf{N_L}$. 

\begin{equation}\label{eq:fullyconstrainR}
\mathbf{R^L_C} \mathbf{N_C}=\mathbf{N_L}
\end{equation}

 Verma et al. \cite{Verma2019} modified this equation to allow more normals to be included in $\mathbf{N_L}$ and $\mathbf{N_C}$ to improve the robustness of the algorithm by overconstraining the problem, hence side-stepping the linear dependency issues. Including more samples in the computation is a common approach in existing methods \cite{Park2014}. However in practice, it was found that while more poses was more robust, it was also prone to overfitting the chessboard. In this work, instead of using more poses for a single calibration result, we chose to take the approach of using multiple sets of 3 poses, to obtain a robust estimate of the calibration parameters with uncertainty. Using 3 poses fully constrains Eq. (\ref{eq:fullyconstrainR}), and also makes $\mathbf{N_C}$ and $\mathbf{N_L}$ a square matrix. Having a square matrix allows for more meaningful analysis of the matrix. Comparing Fig. \ref{fig:k3} and Fig. \ref{fig:k9} shows that a good result can be obtained with sets of 3 and 9 on our baseline method if the correct poses are chosen. 
 
\subsection{LINEAR DEPENDENCE IN THE NORMAL MATRIX}
To avoid redundant poses, it is important to be able to identify linear dependence in the normal matrix as we need to compute $\mathbf{N_C}^{-1}$ when we solve for $\mathbf{R^C_L}$ in Eq. (\ref{eq:fullyconstrainR}). When two or more rows are close to being linearly dependent in the normal matrix, the computation of the inverse becomes unstable and can adversely affect calibration results. This issue is often side-stepped when an expert in calibration knows which poses to use, however for an inexperienced user, it can be challenging to know how to do so.

\begin{figure}[thpb]
  \centering
  \vspace{2mm}
  \includegraphics[width=\columnwidth]{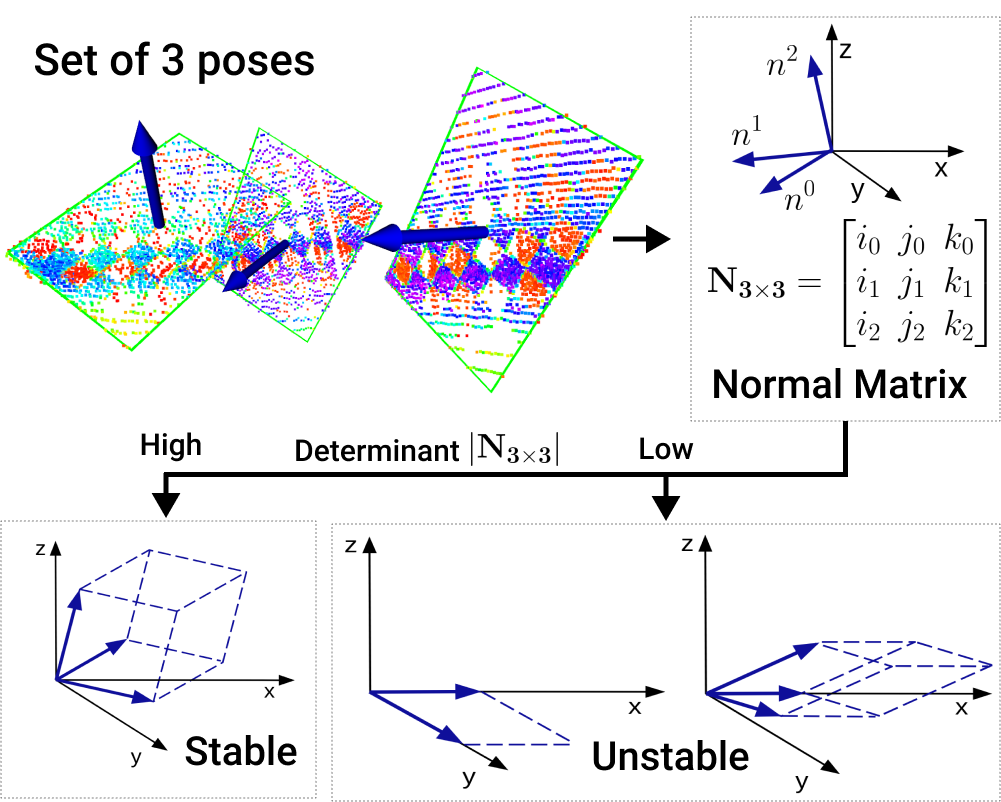}
  \caption{\small Illustration of linear dependency. When the normal (blue arrows) of the target planes have enough variation, it outlines a distinct parallelopiped (stable) with a large determinant, which is the volume of this parallelopiped. However, if the normals have similar or exactly the same orientations, the parallelopiped is hard to define or non-existent, and the determinant is close to zero.}
  \label{fig:determinant}
\end{figure}

The normal matrix $\mathbf{N_{3\times 3}}$ outlines a parallelepiped whose signed volume can be computed by taking the determinant $|\mathbf{N_{3\times 3}}|$ which is the volume scaling factor of the linear transformation of $|\mathbf{N_{3\times 3}}|$. When the determinant is zero, there is little to no definition of the parallelepiped, its volume can't be well computed and hence the computation of its inverse is unstable as illustrated in Fig. \ref{fig:determinant}.  

The linear dependency of the matrix can also be computed by looking at the rank, or the condition number $\kappa$ of the matrix \cite{Zhou2012, Zhou2018}. 

\begin{equation}\label{eq:conditionofmatrix}
\kappa(\mathbf{N}) = ||\mathbf{N^{-1}}\mathbf{N}||
\end{equation}

\begin{equation}\label{eq:avg_cond}
\kappa_{LC} = \max\{\kappa({\mathbf{N_L}}),\kappa({\mathbf{N_C}})\}
\end{equation}

The condition number of the normal matrix $\kappa(\mathbf{N_{3 \times 3}})$, shown in Eq. (\ref{eq:conditionofmatrix}), gives an indication of the stability of its inverse $\mathbf{N_{3 \times 3}^{-1}}$ where most stable result gives $\kappa = 1$ and the least stable result giving a value that grows towards $+\infty$. In theory, the condition number of $\mathbf{N_C}$ should be exactly the same as $\mathbf{N_L}$ as the relative orientation of the poses in the set should be the same whether in pointcloud or image domain. However, this is not always the case and therefore we take the worst case scenario of the two condition numbers for a more robust measure. We compute the condition number of the normal matrices using the frobenius norm to assess linear dependency.

\subsection{ERROR IN LIDAR MEASUREMENTS}
\begin{figure}[thpb]
  \centering
  \includegraphics[width=.7\columnwidth]{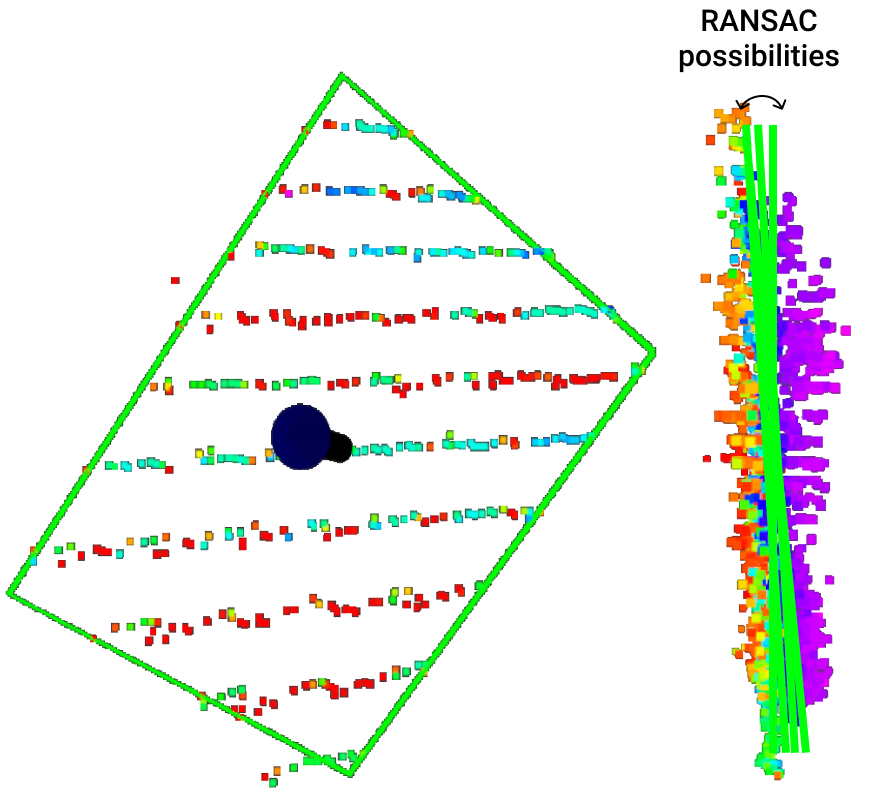}
  \caption{\small Lidar range error may cause it to pick up a skewed board (left), or incorrectly fit a plane to the chessboard (right). Both of these affect the accuracy of feature extraction from the target.}
  \label{fig:lidar_errors}
\end{figure}

To get a good estimation of calibration parameters for target-based methods, an accurate feature extraction of the target is expected. However in reality, lidars and cameras both have errors in measurements that affect the accuracy of the feature extraction process. Lidars tend to have errors in their range accuracy; Velodyne VLP-16 for example documents a range accuracy of $\pm 3$ cm. Our baseline method first determines the edge lines of the target with Random sampling consensus (RANSAC), then finds the intersection of edge lines to identify the corners of the chessboard. This estimation of chessboard corners is heavily influenced by the number of points that lie on the board, and the range accuracy of those points as illustrated in Fig. \ref{fig:lidar_errors}. 

It was found empirically that the errors of chessboard measurements had a greater error at certain distances from the sensor pair (such as being too close or too far), and also when the board faced certain directions. Therefore, to assess the accuracy of sensor measurements, the sum of dimension errors $e_{dim}$ for the chessboard was used as described in Eq. (\ref{eq:dim_sum_error}) where $l_{L,i}$; $i=0,1,2,3$ are the lidar's measurement of the chessboard's edge lengths, and $l_{M,i}$; $i=0,1,2,3$ are the physically measured chessboard edge lengths. For each set of poses, we get 3 errors $e_{dim,j}$; $j=0,1,2$, and the mean error of the 3 poses are taken as the set's average board error $e_{be}$ in Eq. (\ref{eq:dim_avg}) in millimetres.

\begin{equation}\label{eq:dim_sum_error}
e_{dim} = \sum^3_{i=0}|l_{L,i} - l_{M,i}|
\end{equation}

\begin{equation}\label{eq:dim_avg}
e_{be} = \frac{1}{3}\sum^2_{j=0}e_{dim,j}
\end{equation}

By assessing the error of the board measurements, we can have a quantitative measure of the accuracy in chessboard corner and centre estimation. As the translation vector is computed from aligning the centre of the chessboard in the camera and lidar domain, an error in centre estimation will lead to errors in calibration results. 

\subsection{VARIABILITY OF QUALITY (VOQ) METRIC}
For a set of 3 poses, we can assess the quality of the rotation parameters using the condition of the matrix, and the translation parameters using the errors in chessboard measurements. We combine these metrics into a single equation defined in Eq. (\ref{eq:voq}), called Variability of Quality (VOQ). 

\begin{equation}\label{eq:voq}
VOQ = \kappa_{LC} + e_{be}
\end{equation}

A low VOQ score for a set of poses, leads to better generalisation of the calibration parameters to the scene as well as less variability in calibration results. On the other hand, a high VOQ score leads to overfitting and a higher standard deviation of estimated calibration results. 

To estimate the calibration parameters, we propose to use 50 sets of 3 with the lowest VOQ scores out of all possible combinations of the given data. For this reason, our approach thrives when more data is provided. The calibration parameters are calculated by taking the mean of the 50 sets after filtering results that are outside 2 standard deviations from the mean using a z-score. This standard deviation of the resultant sets can also be used as an indication of the certainty of the estimated calibration parameters. 

\begin{figure}[thpb]
  \centering
  \includegraphics[width=.8\columnwidth]{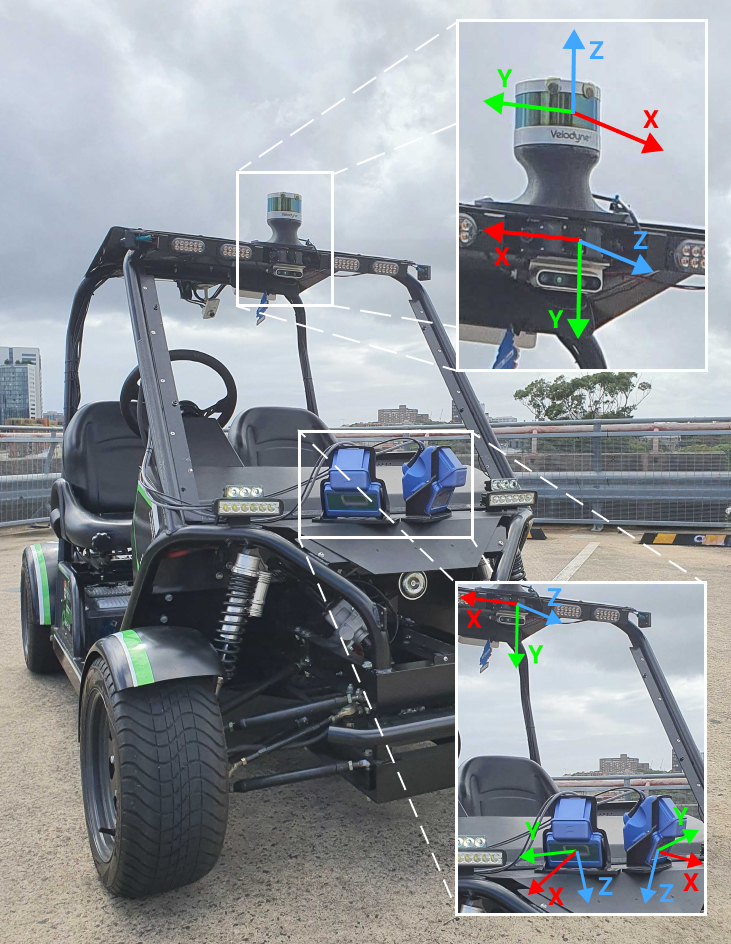}
  \caption{\small Sensor setup. Top-right shows the coordinate frames of the VLP-16 and camera, and the bottom-right shows the frames of the two Baraja Spectrum-Scan™ sensorheads and camera. The Baraja Spectrum-Scan™ sensorheads had to be calibrated with the camera separately. }
  \label{fig:sensor_setup}
\end{figure}

\section{Experimental results}

Experiments were conducted using a chessboard with a 7x5 internal corners, of square length 95mm and board dimensions of $610\times 850$ mm. The sensors used were a Nvidia 2Mega SF3322 gmsl camera, Velodyne VLP-16 and Baraja Spectrum-Scan™ lidar. The VLP-16 has 16 rings with a 360° horizontal and 30° vertical field of view (FOV). The Baraja Spectrum-Scan™ has a 110° horizontal and 25° vertical FOV with adjustable pointcloud resolution, allowing us to focus the points on regions of interest within the FOV. For calibration, we configured the Baraja Spectrum-Scan™ such that the points were always the most dense on the chessboard, which led to lower board dimension errors. We assume that the pre-requisite step of intrinsic camera calibration has been undertaken. The experimental setup is shown in Fig. \ref{fig:sensor_setup}. We compared the brute-force approach of computing the VOQ for every possible combination of poses, with a clustering approach where we selected normals from clusters to minimise the condition number $\kappa$. However as the number of combinations are not very large for sets of 3, we adopted the brute-force approach which was faster. Our proposed calibration pipeline takes around 90s (excluding the manual data collection process).  

\begin{figure}[thpb]
  \centering
  \includegraphics[width=\columnwidth]{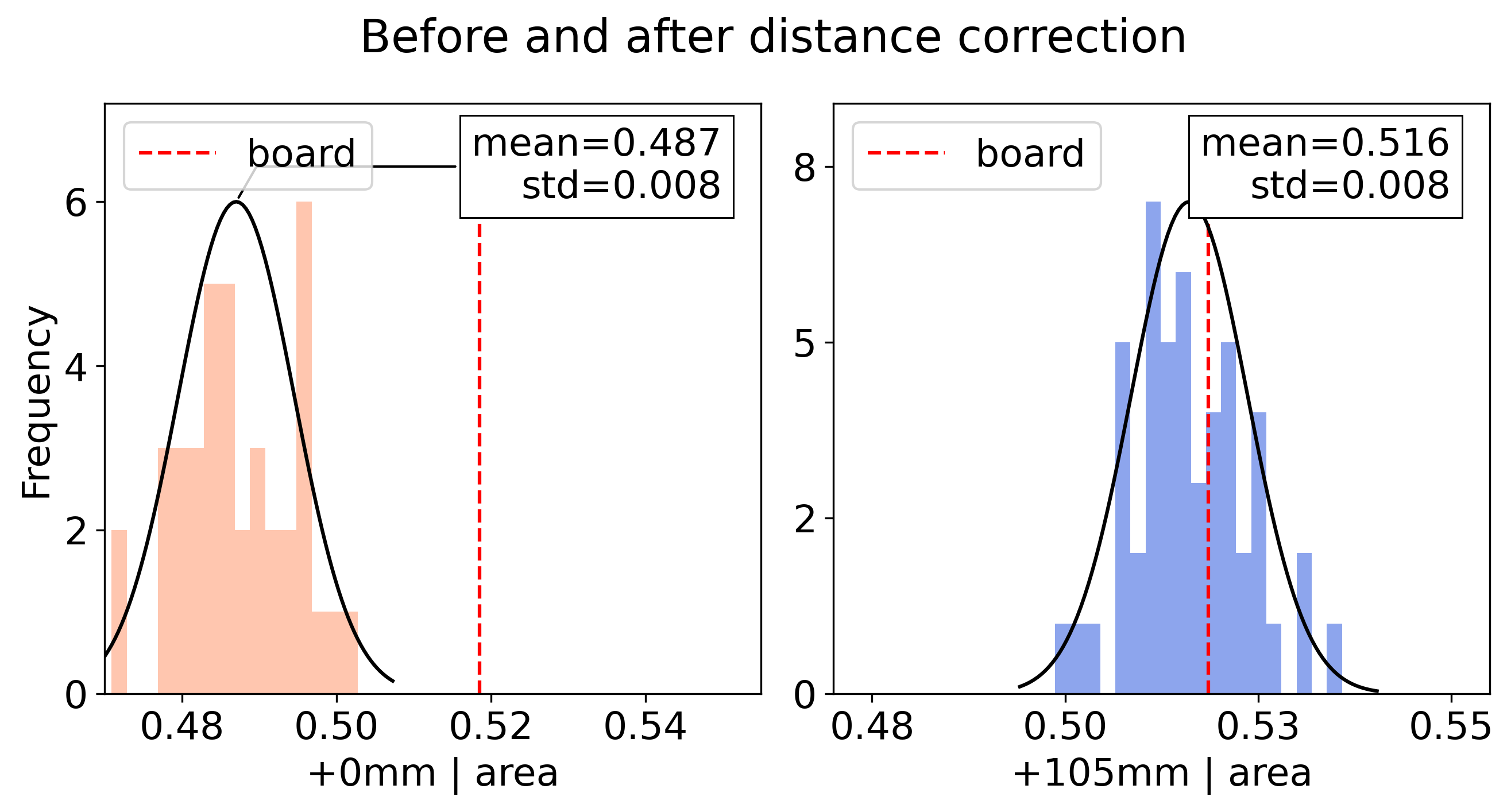}
  \caption{\small Distance offset for the Baraja Spectrum-Scan™ in metres squared. Before the offset was added, the chessboard's area was consistently underestimated on every chessboard sample.}
  \label{fig:distoffset}
\end{figure}

 \begin{figure}[htbp]
    \centering
    \begin{subfigure}[t]{.5\columnwidth}
        \centering
        \includegraphics[width=\columnwidth]{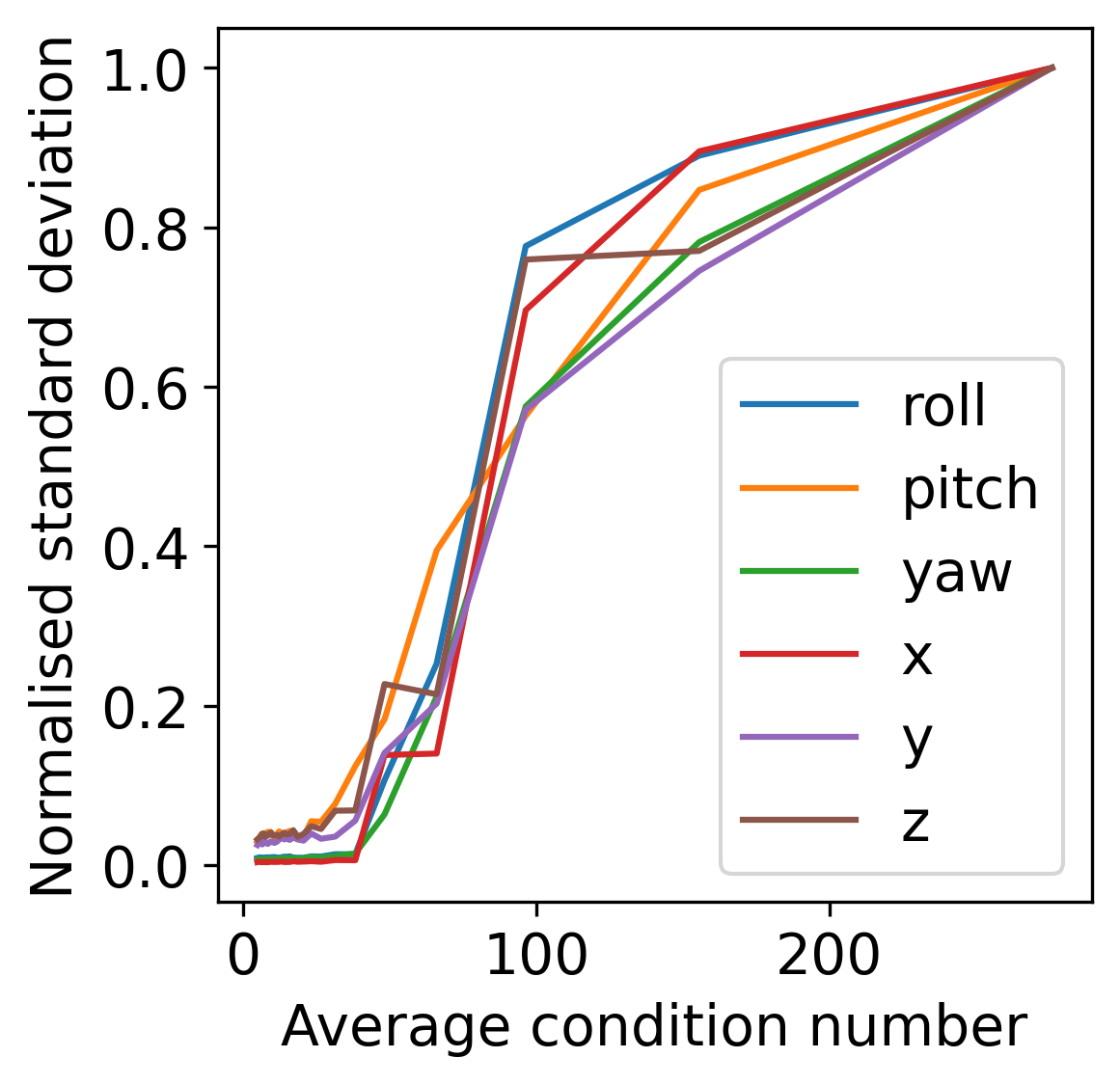}
        \caption{}
        \label{fig:cond_std}
    \end{subfigure}%
    \hfill
    \begin{subfigure}[t]{.5\columnwidth}
        \centering
        \includegraphics[width=\columnwidth]{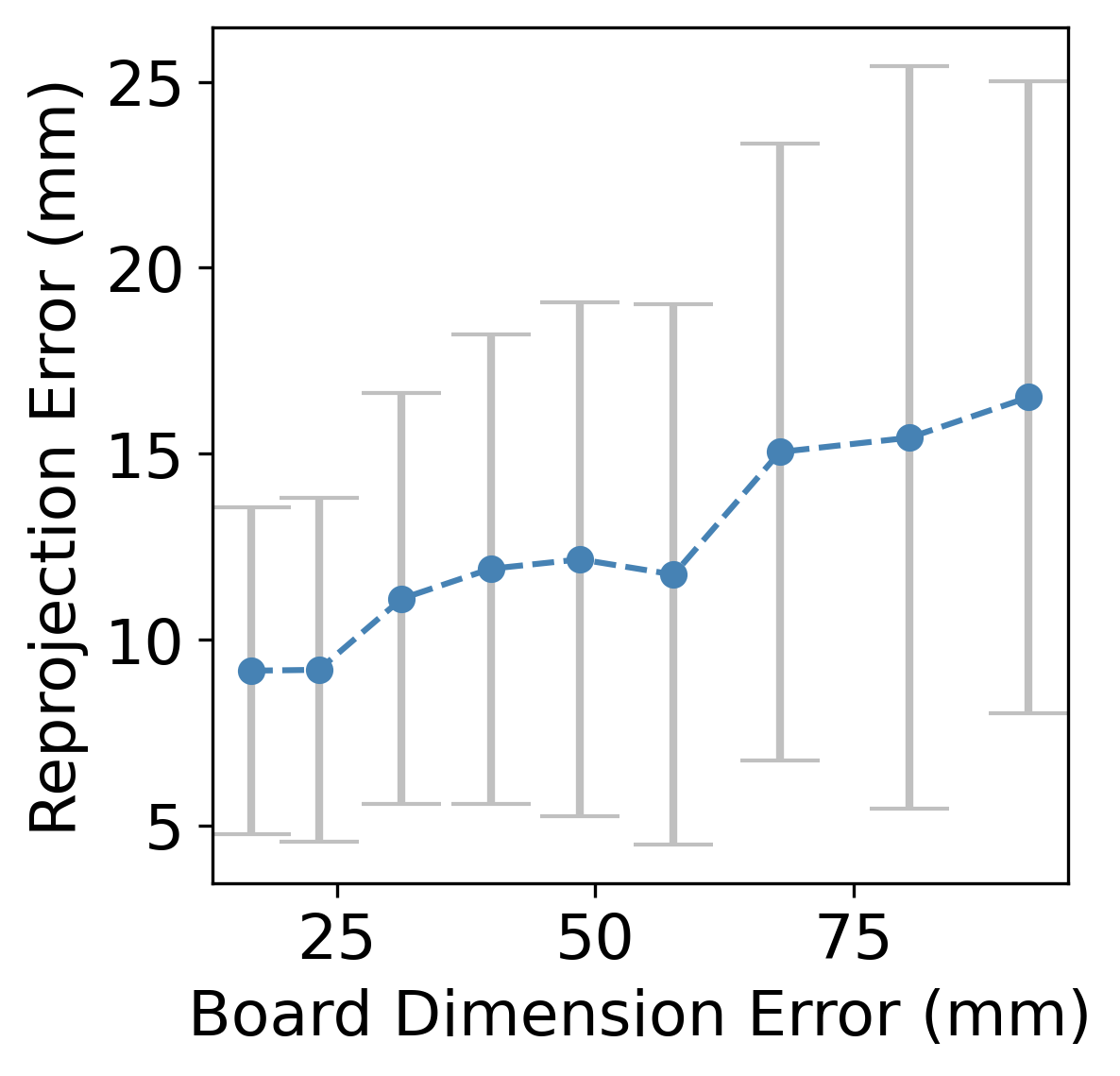}
        \caption{}
        \label{fig:be_repro}
    \end{subfigure}
    \caption{\small Plots for VLP-16 data. (a) Demonstrates the effect of condition number on the stability of calibration results. (b) Shows that a lower chessboard dimension error leads to a lower reprojection error mean and standard deviation.}
\end{figure}

\subsection{DATA COLLECTION}

A total of 50 poses were collected for both pairs of VLP16-camera and Baraja Spectrum-Scan™-camera, which generated ${50 \choose 3} = 19600$ combinations. When capturing these poses, we ensured that there was variation in the pitch and yaw axes of the target. For the roll axis, we kept the chessboard at a 45° angle to form a diamond shape to better detect chessboard edges with a low resolution lidar. We noticed that the sensor measurements of both the VLP-16 and Baraja Spectrum-Scan™ lidars were consistently under/over estimated and therefore converted each point to its polar form and added a distance offset to it as shown in Fig. \ref{fig:distoffset}. 

The VLP-16 and Baraja Spectrum-Scan™ lidars we used required a distance offset of -30mm and +105mm respectively. If the internal distance bias of the lidars are not accounted for, it is hard to obtain a perfect calibration for the sensor pair as all the target measurements will be inaccurate. For each of the 19600 sets of 3 generated, we computed a calibration result with our baseline method, taking around 6 hours to compute with a cpu on a Dell Optiplex7070 desktop.

\begin{figure}[H]
    \centering
    \begin{subfigure}[t]{\columnwidth}
        \centering
        \includegraphics[width=.85\columnwidth]{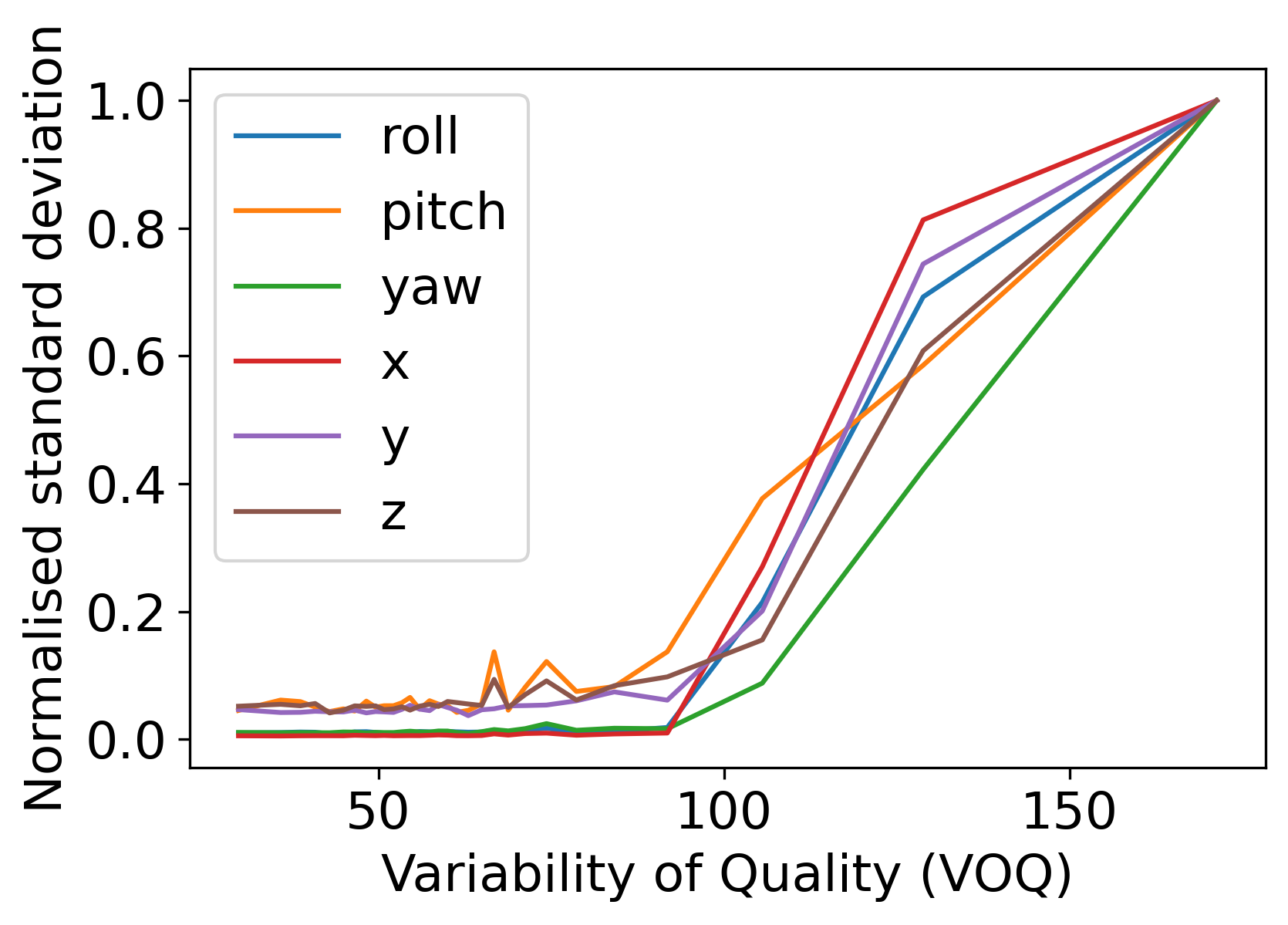}
        \caption{}
        \label{fig:voq_std}
    \end{subfigure}%
    
    \begin{subfigure}[t]{\columnwidth}
        \centering
        \includegraphics[width=.85\columnwidth]{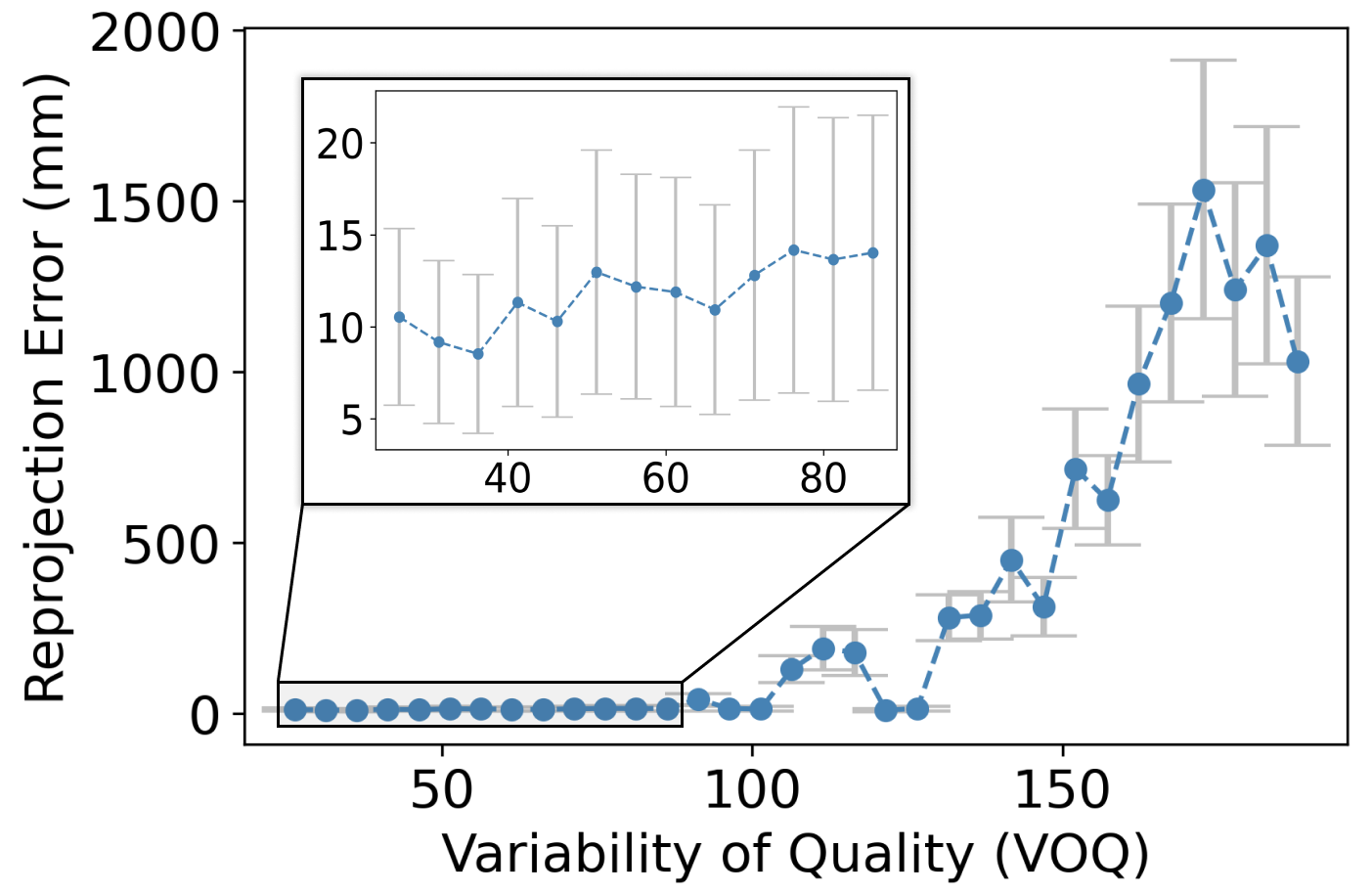}
        \caption{}
        \label{fig:voq_repro}
    \end{subfigure}
    \caption{\small (a) Demonstrates the effect of the VOQ score on the stability of calibration results. (b) Mean reprojection errors for VLP-16 across 46 poses showing that a lower VOQ score leads to lower mean reprojection errors. After VOQ=200, the reprojection errors grow exponentially past 2000 and therefore have been excluded.}
    \label{fig:voq_graphs}
\end{figure}

\subsection{REPROJECTION ERROR FOR ALIGNMENT OF FEATURES ON THE ENTIRE SCENE}
Often when assessing the quality of the calibration parameters, existing methods would project the corners or centre of the target into the image to compute the reprojection error for that particular pose. To assess the efficacy of our metric, we computed the reprojection error over 46 VLP-16 poses that are evenly distributed in the range of 1.7m to 4.5m away from the sensor pair. We did not record poses that were farther away as the low resolution of VLP-16 posed a challenge for RANSAC to accurately outline board edges.  By computing the reprojection error over all 46 poses, we can get a more robust assessment of how well the estimated calibration parameters fit the entire scene, along with the standard deviation of the reprojection error.

\subsection{ASSESSMENT OF METRIC FEATURES}

The stability of the computation of rotation matrix can be gauged using $\kappa_{LC}$ which we assess by grouping the 19600 sets and their corresponding calibration results in equal sized bins. For each bin, the mean $\kappa_{LC}$ and standard deviation of calibration results were computed with results shown in Fig. \ref{fig:cond_std}. Here we observe that the condition number of the normal matrix starts to affect the stability of calibration results at $\kappa_{LC} \approx 50$. Board error was assessed by first filtering out all sets with a condition number higher than 20, then plotting board error with the respective reprojection error as shown in Fig. \ref{fig:be_repro}. We filtered out the condition numbers higher than 20 as they would greatly affect the stability of estimated calibration parameters, and hence affect the assessment of the board dimension error. The smallest average reprojection error achieved on the test dataset was around 1cm. This reprojection error, along with its standard deviation steadily grows as the board dimension errors increase. We provide an easy set of guidelines to facilitate the data collection process in our code repository.

\begin{figure}[H]
     \centering
    \begin{subfigure}[t]{0.49\columnwidth}
        \raisebox{-\height}{\includegraphics[width=\textwidth]{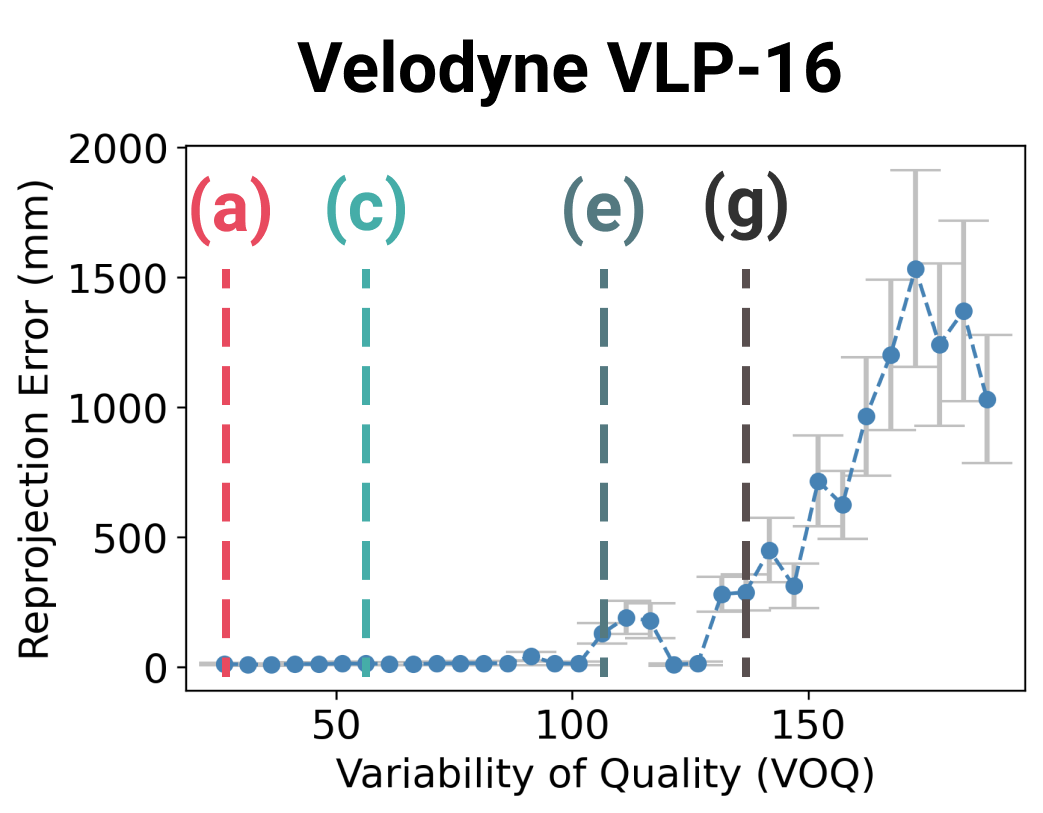}}
    \end{subfigure}
    \hfill
    \begin{subfigure}[t]{0.49\columnwidth}
        \raisebox{-\height}{\includegraphics[width=\textwidth]{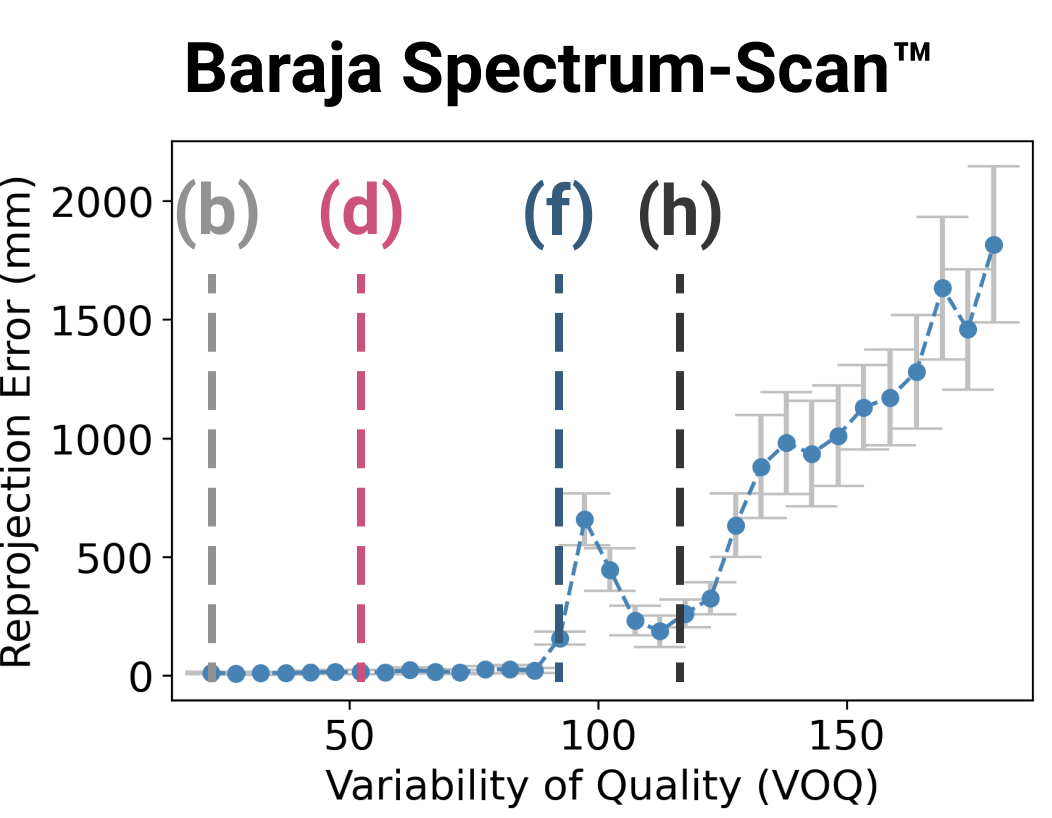}}
    \end{subfigure}
    \par\bigskip
    \begin{subfigure}[t]{0.49\columnwidth}
        {\includegraphics[width=\textwidth]{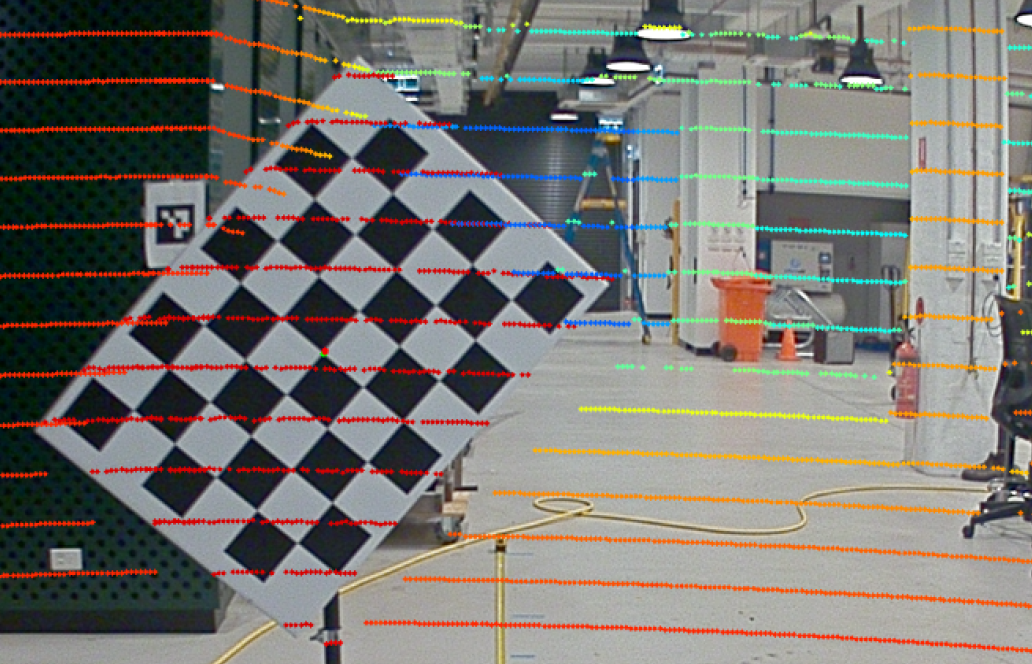}}
        \caption{VOQ = 26.21}
    \end{subfigure}
    \hfill
    \begin{subfigure}[t]{0.49\columnwidth}
        {\includegraphics[width=\textwidth]{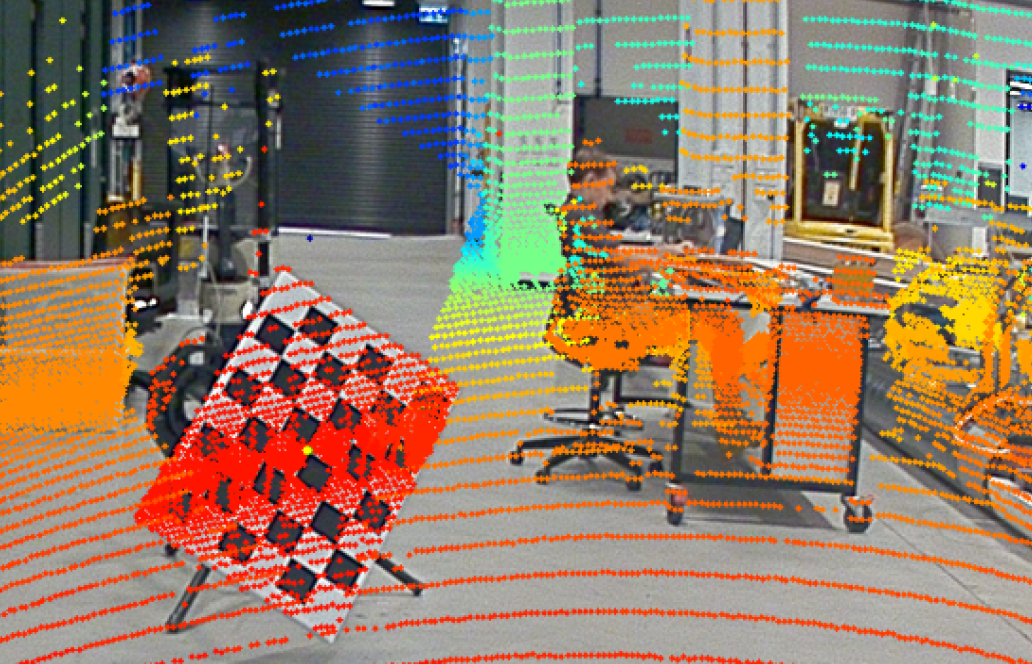}}
    \caption{VOQ = 22.18} 
    \end{subfigure}
    \par\medskip
    \begin{subfigure}[t]{0.49\columnwidth}
        \raisebox{-\height}{\includegraphics[width=\textwidth]{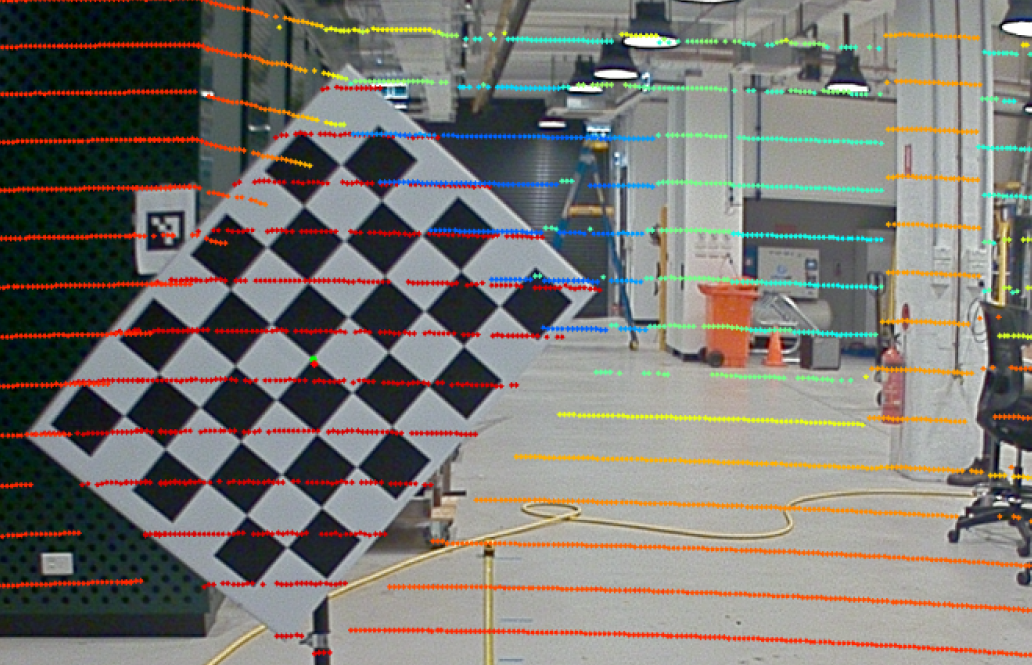}}
        \caption{VOQ = 61.23}
    \end{subfigure}
    \hfill
    \begin{subfigure}[t]{0.49\columnwidth}
        \raisebox{-\height}{\includegraphics[width=\textwidth]{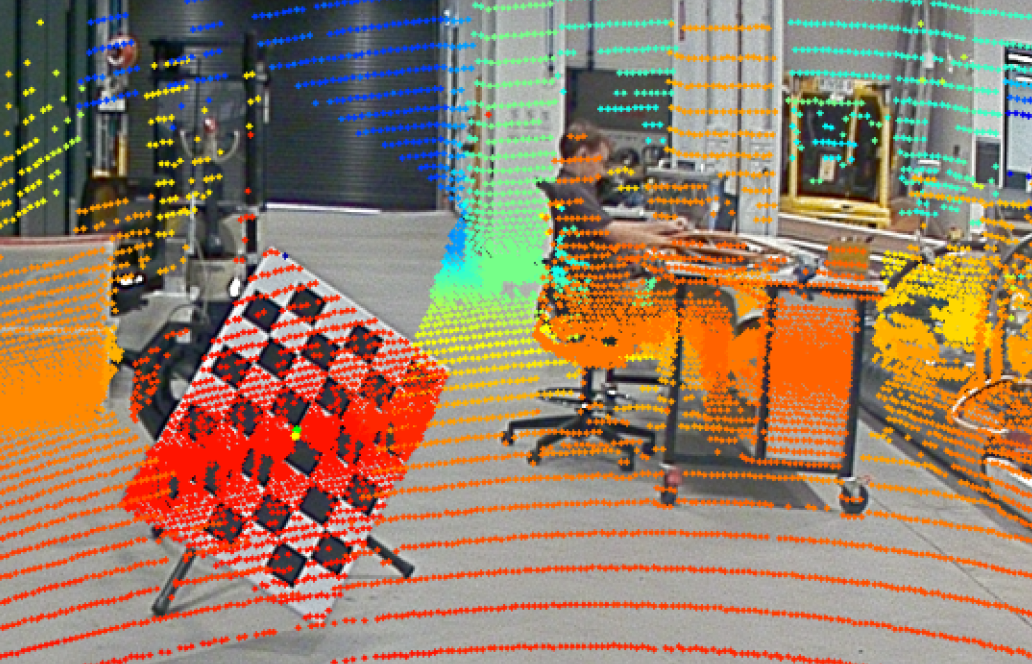}}
        \caption{VOQ = 62.20} 
    \end{subfigure}
    \par\medskip
    \begin{subfigure}[t]{0.49\columnwidth}
        \raisebox{-\height}{\includegraphics[width=\textwidth]{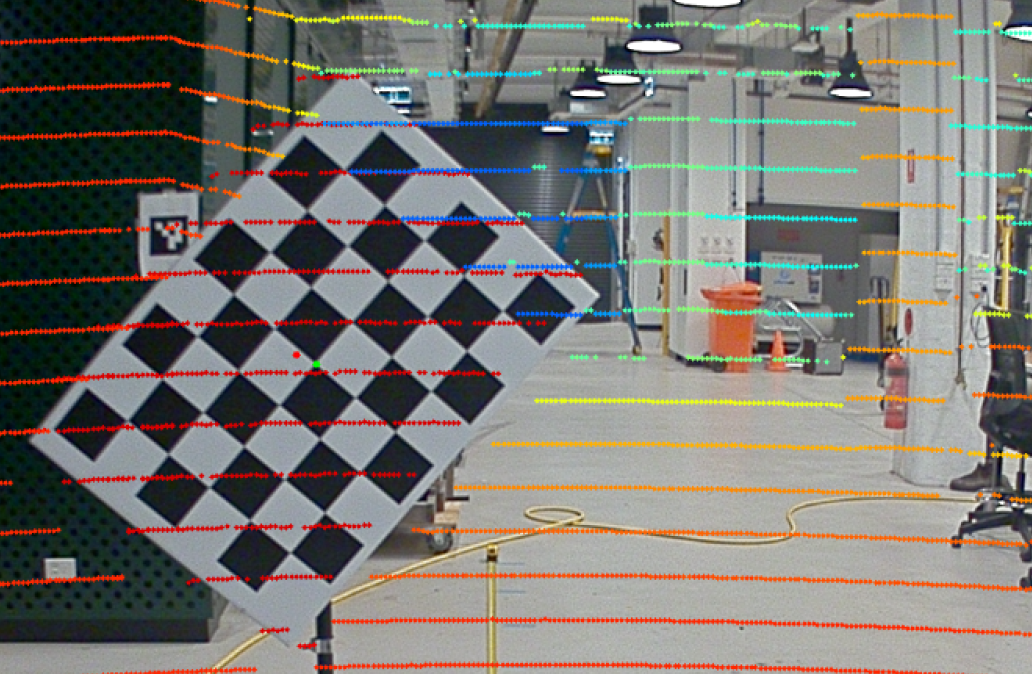}}
        \caption{VOQ = 111.42}
    \end{subfigure}
    \hfill
    \begin{subfigure}[t]{0.49\columnwidth}
        \raisebox{-\height}{\includegraphics[width=\textwidth]{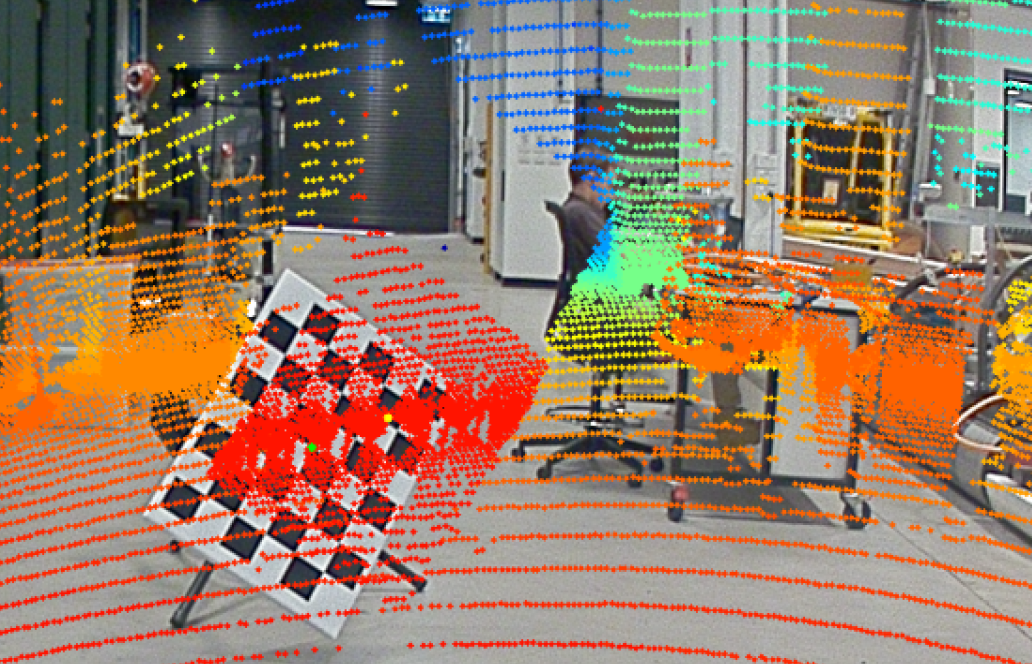}}
        \caption{VOQ = 107.43} 
    \end{subfigure}
    \par\medskip
    \begin{subfigure}[t]{0.49\columnwidth}
        \raisebox{-\height}{\includegraphics[width=\textwidth]{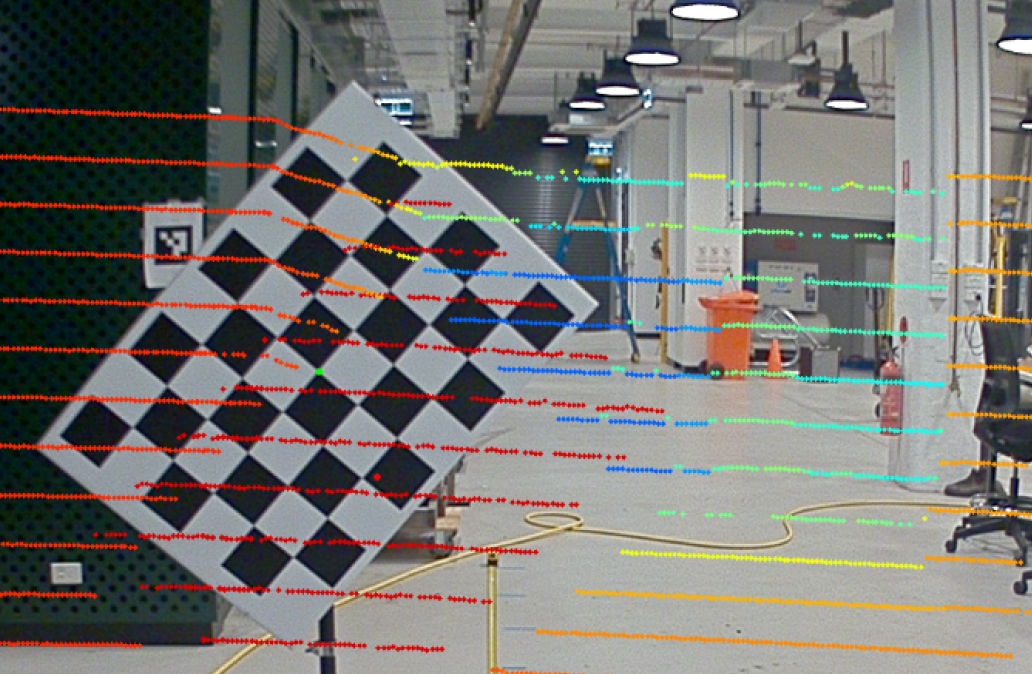}}
        \caption{VOQ = 136.76}
    \end{subfigure}
    \hfill
    \begin{subfigure}[t]{0.49\columnwidth}
        \raisebox{-\height}{\includegraphics[width=\textwidth]{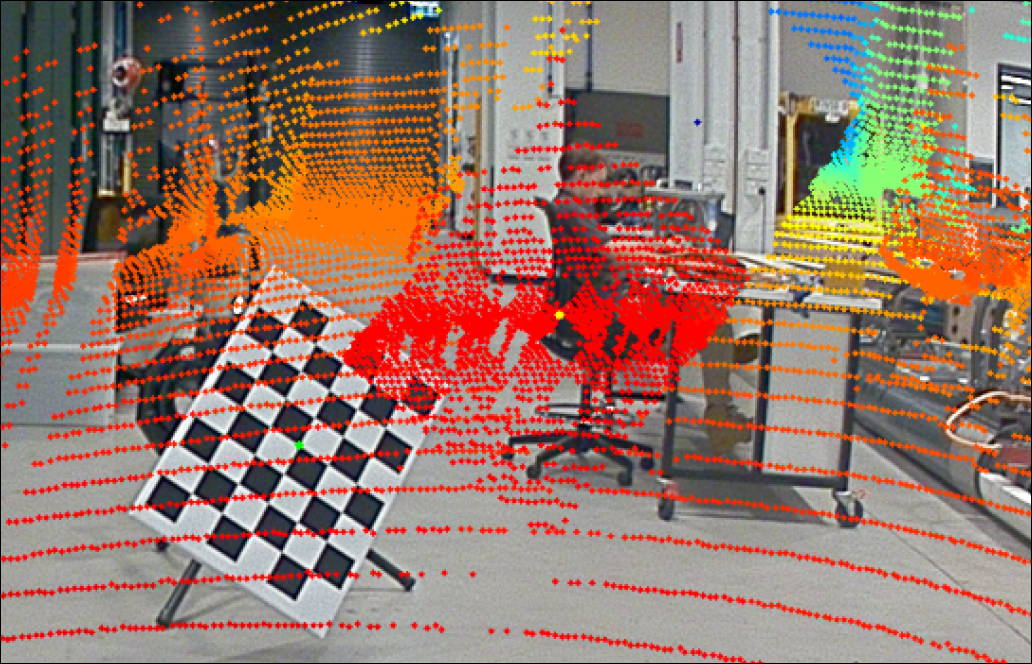}}
        \caption{VOQ = 127.70} 
    \end{subfigure}
    \caption{\small Projection of lidar pointcloud onto image domain using estimated calibration parameters at the VOQ score indicated by the dotted lines. The left side is the calibration of the VLP16-camera pair, and the right is for the Baraja Spectrum-Scan™-camera pair. As the average VOQ score of the 50 sets increases, so too does the scene misalignment, and reprojection error mean and standard deviation. This correlation is observed in both the VLP-16 and Baraja Spectrum-Scan™ lidar.}
    \label{fig:voq_visualisation}
\end{figure}

VOQ was assessed in a similar manner to the condition number and board errors as shown in Fig. \ref{fig:voq_graphs}. In Fig. \ref{fig:voq_std} it can be seen that the uncertainty of calibration results grows as the VOQ score increases. Fig. \ref{fig:voq_repro} shows that an increase in VOQ doesn't simply result in more variability in calibration results, but also a higher average reprojection error of the scene, and a growing standard deviation. For VLP-16 and Baraja Spectrum-Scan™ lidars, the lowest average reprojection errors achieved were both in the range of 1-1.2cm, with a standard deviation of 0.4-0.6cm. In Fig. \ref{fig:voq_visualisation}, the lidar pointclouds for both VLP-16 and Baraja Spectrum-Scan™ have been projected onto the images at specific VOQ scores for a visual comparison to validate the effectiveness of our approach. We emphasise on using both an average reprojection error in combination with a visualisation of the entire scene to assess calibration quality. It can be seen that a lower VOQ score is correlated with a calibration estimation that fits well to the scene. When the VOQ score is slightly increased, it can be seen in the 2nd row that the reprojection error is still low, but the background is not well fitted due to the small calibration errors becoming more noticeable at a distance. As the VOQ score increases, the instability of a high condition number for $\mathbf{N}$ starts to affect the estimated parameters, resulting in an unreliable and highly uncertain estimation. There is a spike in reprojection error at $VOQ \approx 100$ which is caused by the sharp increase in uncertainty of the $\kappa_{LC} \approx 100$. The decrease that follows could be attributed to a mix of sets with high board errors but a low condition number.

\section{CONCLUSION}
We proposed a novel and robust approach to the calibration task that is heavily focused optimising the data selection process. We provide transparency in our results by projecting the entire pointcloud and evaluating our estimated calibration parameters on their ability to fit the entire scene instead of their ability to fit a single pose well. Our method achieves an average reprojection error of 1-1.2cm, with a standard deviation of 0.4-0.5cm, and takes 90s for the whole pipeline. This approach carries with it the added benefit of simplifying the calibration process for practitioners of any calibration expertise, who only needs to focus on providing lots of varied data to the algorithm. Furthermore, we enable the user to quantitatively assess the data that was provided, and also the quality of calibration parameters that is to be estimated. To facilitate the use of these tools we provide the software with a comprehensive installation and use tutorials including examples taken in our lab.

\bibliography{voq_calibration}

\end{document}